\documentclass[letterpaper]{article} 
\usepackage{aaai2026}  
\usepackage{times}  
\usepackage{helvet}  
\usepackage{courier}  
\usepackage[hyphens]{url}  
\usepackage{graphicx} 
\usepackage{natbib}  
\usepackage{caption} 
\usepackage{algorithm}
\usepackage{algorithmic}

\usepackage{booktabs}       
\usepackage{amsfonts}       
\usepackage{amsmath}
\usepackage{nicefrac}       
\usepackage{microtype}      
\usepackage{xcolor}         
\usepackage{graphicx}
\usepackage{algorithm}
\usepackage{algorithmic}
\usepackage{bm}
\usepackage{colortbl}
\usepackage{color}
\usepackage{graphicx} 
\usepackage{multirow}
\usepackage{amsfonts}
\usepackage{dsfont}
\usepackage{mathtools} 

\DeclareMathOperator*{\argmin}{arg\,min}

\usepackage{newfloat}
\usepackage{listings}
\DeclareCaptionStyle{ruled}{labelfont=normalfont,labelsep=colon,strut=off} 
\floatstyle{ruled}
\newfloat{listing}{tb}{lst}{}
\floatname{listing}{Listing}
\title{Auxiliary Gene Learning:\\Spatial Gene Expression Estimation by Auxiliary Gene Selection}
\author{
    Kaito Shiku\textsuperscript{\rm 1}, 
    Kazuya Nishimura\textsuperscript{\rm 2}, 
    Shinnosuke Matsuo\textsuperscript{\rm 1}, 
    Yasuhiro Kojima\textsuperscript{\rm 2}, 
    Ryoma Bise\textsuperscript{\rm 1}
}
\affiliations{
    \textsuperscript{\rm 1} Department of Advanced Information Technology, Kyushu University, Japan\\
    \textsuperscript{\rm 2} Laboratory of Computational Life Science, National Cancer Center Japan\\

    kaito.shiku@human.ait.kyushu-u.ac.jp
}

\usepackage{bibentry}

\begin{document}

\maketitle

\begin{abstract}
Spatial transcriptomics (ST) is a novel technology that enables the observation of gene expression at the resolution of individual spots within pathological tissues.
ST quantifies the expression of tens of thousands of genes in a tissue section; however, heavy observational noise is often introduced during measurement.
In prior studies, to ensure meaningful assessment, both training and evaluation have been restricted to only a small subset of highly variable genes, and genes outside this subset have also been excluded from the training process.
However, since there are likely co-expression relationships between genes, low-expression genes may still contribute to the estimation of the evaluation target.
In this paper, we propose $Auxiliary \ Gene \ Learning$ (AGL) that utilizes the benefit of the ignored genes by reformulating their expression estimation as auxiliary tasks and training them jointly with the primary tasks.
To effectively leverage auxiliary genes, we must select a subset of auxiliary genes that positively influence the prediction of the target genes.
However, this is a challenging optimization problem due to the vast number of possible combinations.
To overcome this challenge, we propose Prior-Knowledge-Based Differentiable Top-$k$ Gene Selection via Bi-level Optimization (DkGSB), a method that ranks genes by leveraging prior knowledge and relaxes the combinatorial selection problem into a differentiable top-$k$ selection problem.
The experiments confirm the effectiveness of incorporating auxiliary genes and show that the proposed method outperforms conventional auxiliary task learning approaches.
\end{abstract}

\begin{links}
    \link{Code}{https://github.com/Shiku-Kaito/AGL}
\end{links}

\section{Introduction}
Spatial transcriptomics (ST) is a novel technology that enables the observation of gene expression at the resolution of individual spots within pathological tissues, playing a crucial role in evaluating disease progression and drug efficacy~\cite{staahl2016visualization}.
While spatial gene expression data obtained through ST provides detailed insights into molecular activity within pathological tissue, its high observational cost has prompted the development of neural network models that aim to estimate spatial gene expression from pathological images~\cite{he2020integrating, pang2021leveraging, Zeng2022SpatialTP, yang2023exemplar,yang2024spatial}.

ST quantifies the expression of tens of thousands of genes in a tissue section. Yet, for most genes, the observed counts are extremely low and often contain dropout noise~\cite{mejia2024enhancing, mejia2023sepal} (zero or near-zero measurements that arise from technical limitations rather than true absence of expression). Because these noisy readings are statistically unreliable, they cannot serve as dependable reference values for benchmarking model performance. To ensure meaningful assessment, earlier studies have therefore restricted both training and evaluation to a narrow set of highly variable genes \cite{zheng2017massively, satija2015spatial,stuart2019comprehensive} whose expression consistently rises above the noise floor and is considered trustworthy. 
In practice, the evaluation target genes consist of only a small subset of the total genes, while the remaining tens of thousands are entirely excluded from performance evaluation~\cite{pang2021leveraging, Zeng2022SpatialTP, chung2024accurate, he2020integrating}.

Although the specific relationships have yet to be thoroughly investigated, many of the genes excluded from training may still share regulatory~\cite{m2022shared} or co-expression patterns~\cite{wang2022region} with the target genes used for evaluation. 
Even when their individual measurements are noisy or sparse, these genes can still provide useful contextual signals that help the model learn richer representations for the evaluation target genes, thereby improving prediction accuracy.

\begin{figure*}
      \centering
        \includegraphics[width=1\linewidth]{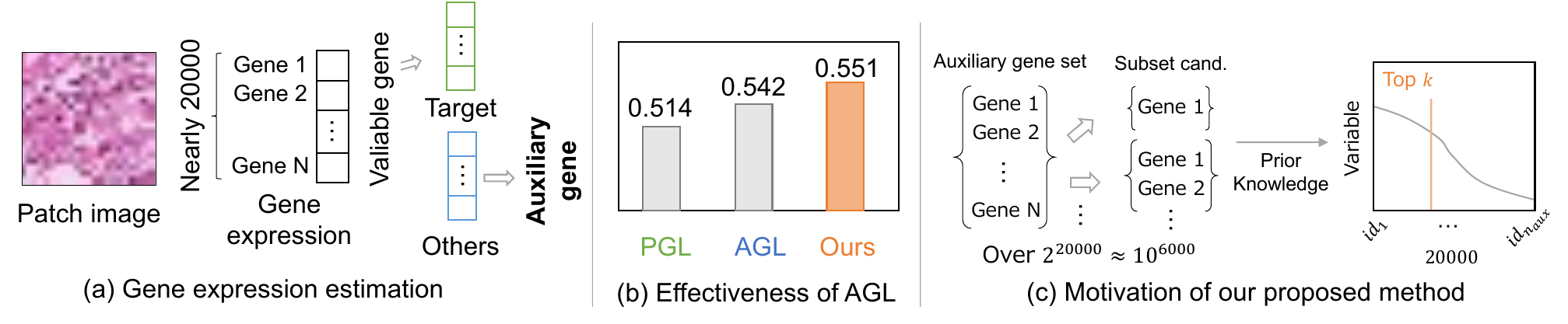}
        \vspace{-6mm}
        \caption{
        (a) Conventional gene expression estimation focuses solely on predicting primary genes, typically ignoring the remaining ones. In this study, we treat these overlooked genes as auxiliary genes. 
        (b) Effectiveness of $Auxiliary \ Gene \ Learning$ (AGL). PGL denotes primary gene learning, which uses only the target gene for training. AGL represents our auxiliary gene learning, which jointly estimates primary genes and previously ignored auxiliary genes, selecting auxiliaries via a differentiable cut-off.
        (c) Illustration of our top-k gene selection approach. As the number of possible subsets exceeds $10^{6000}$, we relax this combinatorial selection into a top-$k$ problem by leveraging prior knowledge of gene-expression signal quality.
        }
        \vspace{-2mm}
        \label{fig:intro}
\end{figure*}

We propose $Auxiliary \ Gene \ Learning$ (AGL) that utilizes the benefit of the ignored genes by reformulating their expression estimation as auxiliary tasks and training them jointly with the primary tasks, as illustrated in Figure~\ref{fig:intro} (a).
An auxiliary task is not evaluated directly, yet learning it together with the primary task can improve the primary-task accuracy (Figure~\ref{fig:intro} (b)). In our case, the primary tasks are the expression predictions for the small subset of highly variable genes, whereas the auxiliary tasks correspond to the remaining genes that were previously discarded.
Introducing this auxiliary task framework into the problem of predicting gene expression from pathological images is, to our knowledge, the first attempt of its kind. Consequently, our study is the first to make practical use of information that earlier work had always thrown away.

However, using all remaining genes as auxiliary targets may be counter-productive, because many of them have very low counts or are dominated by measurement noise. Therefore, we must select a subset of auxiliary genes that positively influence the prediction of the evaluation genes.

As Figure~\ref{fig:intro} (c) shows, nearly twenty thousand genes are available as auxiliary-task candidates. Selecting an appropriate subset from such a vast pool leads to a combinatorial explosion and is far from trivial. Existing work on selective auxiliary-task learning assumes at most five to ten candidate tasks and is designed to pick one or a few from that small set~\cite{kendall2018multi, navon2020auxiliary,ko2023meltr, jiang2023forkmerge, sivasubramanian2023adaptive}. Approaches developed under those conditions do not transfer well to the much larger search space we face here, a limitation that our experiments later confirm.

To overcome this large-scale selection challenge, we introduce Prior-Knowledge-Based Differentiable Top-$k$ Gene Selection via Bi-level Optimization (DkGSB).
DkGSB first ranks all auxiliary-gene candidates by their expression variance, which serves as a simple yet effective proxy for signal quality. Instead of searching over every subset, the method learns a single scalar $k$ that determines how many of the top-ranked genes will remain as auxiliary tasks. A soft and differentiable relaxation makes this top-$k$ operator compatible with gradient descent, and a bi-level objective updates $k$ together with the network weights so that the selected auxiliary genes maximize the accuracy of the target genes.

In experiments conducted on publicly available datasets~\cite{jaume2024hest}, the proposed AGL approach was shown to outperform conventional methods that discard lowly expressed genes.
Furthermore, under the setting where auxiliary genes are used during training, we demonstrate that the proposed method, which incorporates soft parameterization and bi-level optimization with prior knowledge, outperforms conventional auxiliary task learning approaches and also achieves better performance than using all auxiliary genes without selection.

\section{Related work}
\subsection{Gene expression estimation for pathological images}

With recent advances in observation technologies enabling the measurement of gene expression at the resolution of individual spots within pathological tissue, several studies have accordingly explored approaches to estimate gene expression from spot-level pathological images~\cite{he2020integrating, dawood2021all, xie2023spatially, pang2021leveraging, yang2023exemplar, yang2024spatial, chung2024accurate, nishimura2025learning}.
Research on gene expression estimation from spot images has primarily followed two main streams: one in which gene expression is predicted independently for each spot image, and another that incorporates surrounding contextual information into the prediction.
Among the methods that predict gene expression independently for each spot image, the most popular is ST-Net~\cite{he2020integrating}, which employs a CNN-based backbone and formulates gene expression estimation as a multi-output regression task.
Other approaches that have been proposed include stain-aware prediction methods~\cite{dawood2021all} that perform stain deconvolution on H\&E images, and joint embedding methods~\cite{xie2023spatially} that learn a shared representation of spot images and gene expression, similar to CLIP~\cite{radford2021learning}.
In studies that incorporate surrounding context into prediction, methods employing Transformers~\cite{pang2021leveraging}, graph neural networks~\cite{yang2023exemplar,yang2024spatial}, and multi-scale feature embeddings~\cite{chung2024accurate} have achieved state-of-the-art performance by modeling interactions between spots within the WSI, outperforming approaches that treat spot-level predictions independently.

However, these studies commonly train models using only the top-expressing genes with stable expression levels, while discarding low-expressing genes, thereby overlooking potentially informative signals that these discarded genes may provide for learning.
To the best of our knowledge, this is the first attempt to utilize low-expressing genes as auxiliary supervision in the training process.
Furthermore, our proposed auxiliary gene selection method is model-agnostic, making it easily pluggable into existing approaches.

\subsection{Auxiliary task learning}
Auxiliary Task Learning (ATL) aims to improve the performance of a primary task by leveraging information from related auxiliary tasks.
In ATL, various methods have been proposed to select beneficial tasks from multiple auxiliary tasks, such as weighting based on gradient similarity~\cite{chen2018gradnorm}, task-wise uncertainty~\cite{kendall2018multi}, and techniques like ForkMerge~\cite{jiang2023forkmerge}, which create a branch for each auxiliary task combined with the primary task, and aggregate the model parameters through weighted averaging based on the performance improvement rate on the primary task.
In recent years, methods that aggregate auxiliary task losses have become mainstream in the field, including approaches that assign loss weights using learnable parameters~\cite{sivasubramanian2023adaptive}, methods that integrate losses non-linearly using MLPs~\cite{navon2020auxiliary}, and techniques that leverage Transformers to combine losses~\cite{ko2023meltr}.

In typical ATL settings, the number of auxiliary tasks is assumed to range from 1 to 10, or up to around 300 in large-scale cases~\cite{navon2020auxiliary}. In contrast, our proposed auxiliary gene learning (AGL) involves approximately 20,000 auxiliary tasks, making the selection of optimal auxiliary tasks significantly difficult.
Moreover, due to computational cost constraints, methods such as training on all pairs of the primary task and each auxiliary task~\cite{jiang2023forkmerge}, or using Transformers whose computational complexity increases exponentially with the number of tokens~\cite{ko2023meltr}, are practically infeasible to apply to AGL.

\section{Preliminary: general formulation of subset auxiliary task selection}
We first summarize the general formulation of subset selection.
The aim is to choose, from $n_{\mathrm{aux}}$ auxiliary-task candidates, the subset that most improves the performance of the primary tasks evaluated at test time.
The problem can be written as:
\begin{equation}
\label{eq:auxGeneral}
\begin{aligned}
\bm{\lambda}^{\star}
    &= \argmin_{\bm{\lambda}\in\{0,1\}^{n_{\mathrm{aux}}}}
       \sum_{j=1}^{n_{\mathrm{pri}}}
       L_{\mathrm{pri},j}\bigl(\theta^{\star}(\bm{\lambda})\bigr), \\[4pt]
\text{s.t.}\quad
\theta^{\star}(\bm{\lambda})
    &= \argmin_{\theta}\Bigl[
          \sum_{j=1}^{n_{\mathrm{pri}}} L_{\mathrm{pri},j}(\theta)
          + \sum_{j=1}^{n_{\mathrm{aux}}} \lambda_j\,L_{\mathrm{aux},j}(\theta)
       \Bigr],
\end{aligned}
\end{equation}
where \(L_{\mathrm{pri},j}\) and \(L_{\mathrm{aux},j}\) denote the losses for the \(j\)-th primary and auxiliary task, respectively; \(n_{\mathrm{pri}}\) and \(n_{\mathrm{aux}}\) are the numbers of primary and auxiliary tasks.
The binary mask \(\bm{\lambda}\in\{0,1\}^{n_{\mathrm{aux}}}\) specifies which auxiliary genes are selected \((\lambda_j=1)\) or discarded \((\lambda_j=0)\).

However, as noted in the introduction, the number of auxiliary tasks in our setting is enormous ($n_{\mathrm{aux}}\approx 20{,}000$).
The corresponding search space $\binom{n_{\mathrm{aux}}}{k}$ grows exponentially, so solving Eq.~\eqref{eq:auxGeneral} exhaustively is infeasible.

\begin{figure}
\vspace{-6mm}
      \centering        \includegraphics[width=1\linewidth]{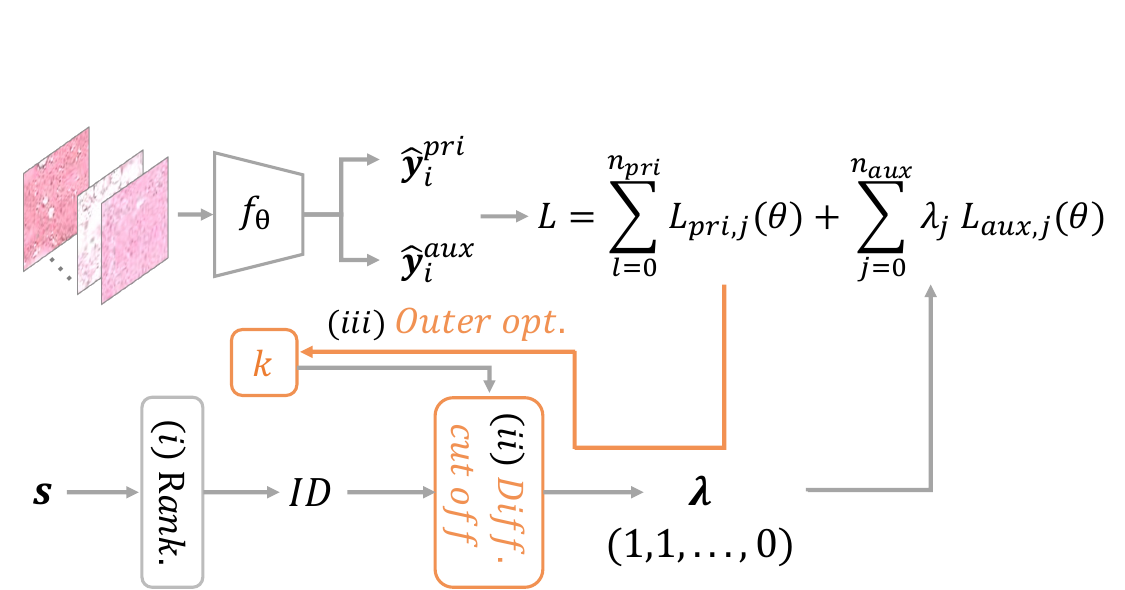}
        \caption{
        Overview of Proposed {\bf DkGSB}. 
        The procedure has three steps:  
(i) auxiliary genes are ranked based on a variance-based score $\bm{s}$;  
(ii) a single learnable scalar $k$ defines a soft top-$k$ mask $\bm{\lambda}(k)$, obtained through a differentiable relaxation of the hard cut-off;  
(iii) $k$ is optimized together with the network weights by a bi-level scheme.  
        }
        \label{fig:overview}
\end{figure}

\section{Auxiliary Gene Learning with Differentiable Top-$k$ Gene Selection}
Figure~\ref{fig:overview} outlines our $Auxiliary \ Gene \ Learning$ (AGL) framework with the Prior-Knowledge-Based Differentiable Top-$k$ Gene Selection (DkGSB) module.  
The procedure has three steps:  
(i) all $\;n_{\mathrm{aux}}\!\approx\!20{,}000$ auxiliary genes are ranked once by a variance-based score (Section 4.2);  
(ii) a single learnable scalar $k$ defines a soft top-$k$ mask $\bm{\lambda}(k)$, obtained through a differentiable relaxation of the hard cut-off;  
(iii) $k$ is optimized together with the network weights by a bi-level scheme (Section 4.3).  
This reduces the combinatorial search to a one-dimensional optimization and retains only auxiliaries that improve primary gene prediction.

\subsection{Problem setup}
Let   
$\mathcal D=\bigl\{(\mathbf{x}_i,\mathbf{y}_i^{\mathrm{pri}},\mathbf{y}_i^{\mathrm{aux}})\bigr\}_{i=1}^{N}$ be the spatial gene expression dataset, where $\mathbf{x}_i$ is the $i$-th spot image, $\mathbf{y}_i^{\mathrm{pri}}\in\mathbb{R}^{n_{\mathrm{pri}}}$ contains the expressions of the $n_{\mathrm{pri}}$ \emph{primary} genes, and $\mathbf{y}_i^{\mathrm{aux}}\in\mathbb{R}^{n_{\mathrm{aux}}}$ stores the remaining $n_{\mathrm{aux}}\approx20{,}000$ \emph{auxiliary} genes. 
Note that $n_{\mathrm{pri}} \ll n_{\mathrm{aux}}$ in this setting.
A neural network $f_{\theta}$ predicts both sets of expressions from an image, $(\hat{\mathbf{y}}_i^{\mathrm{pri}},\hat{\mathbf{y}}_i^{\mathrm{aux}})  = f_{\theta}(\mathbf{x}_i)$.

After ranking, only the top $k$ auxiliary genes are kept; the soft mask $\bm{\lambda}(k)\in[0,1]^{n_{\mathrm{aux}}}$ returned by the relaxation assigns high weights to genes with rank $\le k$ and small weights otherwise. During training, the network weights $\theta$ are updated with losses from both the primary genes and the masked auxiliary genes, while the scalar $k$ is adjusted so that the validation loss of the primary genes decreases. The formal bi-level objective is given in Section 4.3.

\subsection{Prior knowledge-based auxiliary gene ranking} \label{sec:ranking}
In gene expression analysis, it is widely accepted that genes exhibiting high variability relative to their mean expression within biological tissues provide more informative signals~\cite{satija2015spatial}.
We therefore rank all auxiliary genes by a \emph{highly variable gene} (HVG) score, which corrects for the dependence of dispersion on mean expression.


To compute the HVG score, we calculate the mean expression $\mu_j$ and the raw dispersion $\delta_j$ for the $j$-th gene across the entire dataset, where the raw dispersion $\delta_j$ is obtained by dividing the variance by the mean expression.
All genes are divided into twenty bins based on their mean expression $\mu_j$, and within each bin, the dispersions are $z$-score normalized to yield scale-independent HVG scores
\begin{equation}
s_j = \mathrm{z\text{-}score}\!\bigl(\delta_j\bigr),
\qquad
j = 1,\dots , n_{\mathrm{aux}} .
\end{equation}

We then sort the genes in descending order of their HVG scores,
\begin{equation}
\textit{ID} = (\,id_1, id_2, \dots , id_{n_{\mathrm{aux}}}\,), \\
s_{id_1} > s_{id_2} > \dots > s_{id_{n_{\mathrm{aux}}}},
\end{equation}
and pass this ranked list to the differentiable top-\(k\) selection module described in the next subsection.

\begin{figure}
\begin{algorithm}[H]
\caption{Prior Knowledge-Based Differentiable Top-$k$ Gene Selection via Bi-level Optimization.}
\label{alg:algMeta}
\small
  \begin{algorithmic}[1]
    \REQUIRE Training data $\mathcal D^{t}$,  Validation data $\mathcal D^{v}$, Temperature $\tau$, Learning rates  $(\alpha, \beta)$, Inner steps $H$.
    \WHILE{not Converge}
        \WHILE{not Converge}
            \STATE  $\{ \mathbf{x}_i, \mathbf{y}_i \}$ 
            $\leftarrow$ $SampleMiniBatch (\mathcal{D}^{t})$ \\
            \STATE  Obtain $\tilde{\bm{\lambda}}(k,\tau)$ by  Eq.(~\ref{eq:diffLambda}).\\
            \STATE $(\hat{\mathbf{y}}_i^{\mathrm{pri}},\hat{\mathbf{y}}_i^{\mathrm{aux}})  = f_{\theta}(\mathbf{x}_i)$ \\
            \STATE Update: $\theta \xleftarrow{}  \theta - \alpha \nabla_{\theta} \bigl(\sum_{j=1}^{n_{\mathrm{pri}}} L_{\mathrm{pri},j}(\theta)
          + \sum_{j=1}^{n_{\mathrm{aux}}} \tilde{\lambda}_j(k)\,L_{\mathrm{aux},j}(\theta)\bigr)$
       \ENDWHILE
    
        \WHILE{not Converge}
            \STATE  $\{ \mathbf{x}_i, \mathbf{y}_i \}$ $\leftarrow$ $SampleMiniBatch (\mathcal{D}^{t})$ \\
            \STATE  Obtain $\tilde{\bm{\lambda}}(k,\tau)$ by  Eq.(~\ref{eq:diffLambda}).\\
            \FOR {$h = 1, ..., H$}
                \STATE Update: $\theta^+ \xleftarrow{}  \theta - \alpha \nabla_{\theta} \bigl(\sum_{j=1}^{n_{\mathrm{pri}}} L_{\mathrm{pri},j}(\theta)
          + \sum_{j=1}^{n_{\mathrm{aux}}} \tilde{\lambda}_j(k)\,L_{\mathrm{aux},j}(\theta)\bigr)$
            \ENDFOR
            \STATE  $\{ \mathbf{x}_i, \mathbf{y}_i \}$ $\leftarrow$ $SampleMiniBatch (\mathcal{D}^{v})$ \\
            \STATE Update: $k \xleftarrow{}  k - \beta \nabla_{k} \sum_{j=1}^{n_{\mathrm{pri}}} L_{\mathrm{pri},j}(\theta^+)$
       \ENDWHILE
   \ENDWHILE
    
  \end{algorithmic}
\end{algorithm} 
        \vspace{-7mm}
\end{figure}

\subsection{Differentiable top-$k$ selection via bi-level optimization}
\label{sec:bi-level}
Based on gene ranking results, we find the optimal cut-off $k$ that selects the top-ranked auxiliary genes most helpful for predicting the primary genes. The ranking itself is fixed by the HVG scores; only $k$ is optimized. Our AGL optimization is reformulated as:
\begin{equation}
\begin{aligned}
\argmin_{k\in\{1,\dots,n_{\mathrm{aux}}\}}
       &\sum_{j=1}^{n_{\mathrm{pri}}}
       L_{\mathrm{pri},j}\bigl(\theta^{\star}(k)\bigr),
       \quad
\\ \text{s.t.}\quad
\theta^{\star}(k)
    = \argmin_{\theta}\Bigl[
          &\sum_{j=1}^{n_{\mathrm{pri}}} L_{\mathrm{pri},j}(\theta)
          + \sum_{j=1}^{n_{\mathrm{aux}}} \lambda_j(k)\,L_{\mathrm{aux},j}(\theta)
       \Bigr],
\end{aligned}
\end{equation}
where $\lambda_j$ is $1$ if $id_j \le k$ and $0$ otherwise.

In this study, we use the Pearson correlation coefficient loss function for $L_\mathrm{pri}, L_\mathrm{aux}$, which is known to reduce batch-effect–related scaling bias.
\begin{equation}
\label{eq:auxlearning}
\begin{split}
&L = 1 - \frac{\sum_{i=1}^{N} (\hat{y}_i - \hat{Y})(y_i - Y)}{\sqrt{\sum_{i=1}^{N} (\hat{y}_i - \hat{Y})^2} \sqrt{\sum_{i=1}^{N} (y_i - Y)^2}},
\end{split}
\end{equation}
where $N$ denotes the mini-batch~\footnote{In this paper, we refer to a patient as a ``batch,'' and a subset of data used for a single model-update step as a ``mini-batch.''} size, and $Y$ and $\hat{Y}$ represent the mean ground-truth and predicted expression values within the mini-batch, respectively.

Because the ordinary top-$k$ operator is not differentiable, we replace it with a temperature-controlled soft mask so that the cut-off can be updated by gradient descent. The resulting mask $\tilde{\bm{\lambda}}(k,\tau)$ is given by
\begin{equation}
\label{eq:diffLambda}
\tilde{\lambda}_j
  = \frac{\exp(k/\tau)}{\exp(id_j/\tau) + \exp(k/\tau)},
  \qquad j=1,\dots ,n_{\mathrm{aux}},
\end{equation}
where $id_j$ is the rank of the $j$-th auxiliary gene, $k$ is the learnable threshold, and $\tau$ controls the softness of the step. As $\tau\to0$, the mask approaches the hard top-$k$ selection $\mathds{1}[\,id_j\le k\,]$.


Algorithm~\ref{alg:algMeta} learns the network weights $\theta$ and the cut-off $k$ in a bi-level manner. At the start of each outer iteration, $k$ is fixed and an inner loop runs on the training data $\mathcal{D}^{t}$. For each training mini-batch, the current $k$ is converted into a soft mask $\tilde{\bm{\lambda}}(k,\tau)$, the mini-batch is forward-propagated, and $\theta$ is updated with the total loss that combines the primary terms with the auxiliary terms weighted by $\tilde{\bm{\lambda}}$. This sequence is repeated a fixed number of times, denoted $H$, so that $\theta$ is fitted to the current auxiliary subset.

After these $H$ weight updates, a validation mini-batch from $\mathcal{D}^{v}$ is used to refine $k$. The primary-gene validation loss is back-propagated, and a single
gradient step is taken on $k$ through the differentiable mask $\tilde{\bm{\lambda}}$. Because $\tilde{\bm{\lambda}}$ is a smooth function of $k$, standard optimizers with learning rate $\beta$ can be applied directly. By alternating training-data updates of $\theta$ ($H$ steps) with validation-data updates of $k$ (one step), the algorithm converges to the value of $k$ that minimises the primary-gene validation error, reducing an otherwise combinatorial subset search to a one-dimensional
optimization learned end-to-end with the model.

{
\renewcommand{\arraystretch}{1.2}
\begin{table*}[t]
    \def\@captype{table}
      \makeatother
        \centering
        \caption{{\bf Comparison with conventional method}  includes both the baseline setting without auxiliary genes and comparisons with existing auxiliary task learning methods under the ``{\bf AGL}'' setting. The performance of each method is evaluated through cross-validation, and the reported values represent the mean and standard deviation. Best performances are bold.}
        \scalebox{1.0}{
     \begin{tabular}{cc||c|c|c||c||c} \hline
     \multicolumn{2}{c||}{\multirow{2}{*}{Method}} & \multicolumn{3}{c||}{Intra-batch} &Inter-batch & \multirow{2}{*}{\bf Average}   \\ 
          &&{\bf BOWEL A} &{\bf BOWEL B}&{\bf OVARY} &{\bf HEART}&  \\ \hline 
                PGL &&0.514$\pm$0.009&0.419$\pm$0.004&0.448$\pm$0.008&0.245$\pm$0.034&0.407\\
                {\bf AGL} &$+$ ALL&0.542$\pm$0.008&0.430$\pm$0.006&0.451$\pm$0.005&0.248$\pm$0.038&0.418\\
                {\bf AGL}&$+$ Uncertainty &0.543$\pm$0.009&0.431$\pm$0.005&0.451$\pm$0.006&0.252$\pm$0.038&0.419\\
                {\bf AGL}&$+$ AuxLearn &0.541$\pm$0.007&0.411$\pm$0.012&0.445$\pm$0.008&0.251$\pm$0.037&0.412\\
                {\bf AGL}&$+$ AMAL &0.535$\pm$0.010&0.416$\pm$0.005&0.443$\pm$0.007&0.248$\pm$0.039&0.411\\
        \rowcolor{gray!15} {\bf AGL}& $+$ {\bf DkGSB} &{\bf 0.551}$\pm$0.009&{\bf 0.440}$\pm$0.008&{\bf 0.458}$\pm$0.006&{\bf 0.256}$\pm$0.039&{\bf 0.426}\\ \hline 


        \hline  
        \end{tabular}
        }
        \label{tab:comparison_intra}
\end{table*}
}

\section{Experiments}
Spatial gene expression data are frequently affected by batch effects, which arise from differences in experimental conditions or technical variation introduced by the operator during data acquisition. These effects often lead to significant biases in gene expression measurements across different slides~\cite{lopez2018deep, shaham2017removal}.
In this experiment, our objective is to analyze the impact of auxiliary genes on the estimation performance of target genes; however, in the presence of batch effects, there is a risk that the effects of auxiliary genes may not be accurately evaluated.
Therefore, we conduct experiments based on two settings: an intra-batch experiment, where batch effects are absent, and an inter-batch experiment, where training and evaluation are performed across different batches.

\subsection{Dataset}
For both intra-batch and inter-batch experimental settings, we used data from the Hest-1k dataset~\cite{jaume2024hest}, a large-scale public dataset comprising paired spatial gene expression and pathological images.
The details of each experimental setting are as follows.

\noindent
{\bf Intra-batch experiment.}
We used slides from two bowels ({\bf BOWEL A}, {\bf BOWEL B}) and one ovary ({\bf OVARY}) organs, along with spatial gene expression data acquired using Visium technology~\cite{williams2022introduction}.
After quality control, {\bf BOWEL A}, {\bf BOWEL B}, and {\bf OVARY} contain 4{,}096, 4{,}617, and 5{,}774 patch images of size 224×224, respectively, each cropped at the center of a spot with a width of 55 $\mu\mathrm{m}$. The corresponding expression data include 18{,}066, 18{,}054, and 18{,}043 gene types, respectively.
As a preprocessing step, the expression values for each spot were log-normalized.
Following the experimental setting described in the Hest-1k dataset paper and other studies using this dataset~\cite{jaume2024hest, cho2025mv_}, we also use the top 50 highly variable genes as prediction targets.
We split the patch images into five folds using a 3:1:1 ratio for the training, validation, and test sets on each slide, and performed 5-fold cross-validation.\footnote{No data leakage occurs when updating the parameter $k$, as the validation set used is completely independent of the test set.}

\noindent
{\bf Inter-batch experiment.}
We used four slides from the heart organ (\textbf{HEART}), along with spatial gene expression data acquired using Visium technology. Following~\cite{xie2023spatially}, we performed leave-one-batch-out cross-validation by splitting the data into training and evaluation sets based on batch identity.
After quality control, each slide contained 2{,}857, 3{,}400, 3{,}236, and 3{,}042 patch images of size 224×224, respectively. We selected genes of the 15{,}904 type that were commonly present across all batches as training targets.

\subsection{Implementation details and evaluation metric}
We implemented the proposed method using PyTorch~\cite{Paszke2019PyTorchAI}. 
As the feature extractor for $f_{\theta}$, we employed a ResNet18~\cite{he2016deep} model pretrained on ImageNet~\cite{deng2009imagenet}.
To train the network, we used the Adam optimizer~\cite{adam} with a learning rate of $\alpha = 3 \times 10^{-5}$, a mini-batch size of 128, and trained the model for up to 1{,}000 epochs with early stopping set to 20 epochs.
In the bi-level optimization, the number of inner steps was set to $H = 2$, and the learning rate $\beta$ was set to $3 \times 10^{-3}$.
The temperature parameter $\tau$ of the differentiable cut-off is set to 0.01.
The proposed method is trained on a system equipped with an Intel Xeon Gold 5122 CPU and an NVIDIA RTX A6000 GPU.

We evaluated the performance of the proposed method based on the Pearson correlation coefficient ({\bf PCC})~\cite{chung2024accurate}, which is calculated as the correlation between the predicted and ground-truth gene expression values.
The reported value is the average computed over all target genes after calculating the PCC for each gene within the test dataset.



\subsection{Comparison}
To demonstrate the effectiveness of our $Auxiliary \ Gene \ Learning$ ({\bf AGL}) framework, we compared the performance of the following methods:
1) ``Primary gene learning (PGL)'' trains the model using only the primary genes, without incorporating any auxiliary genes.
2) ``{\bf AGL}+ All'' trains the model under the proposed AGL setting, but without any auxiliary–task selection: every gene other than the primary is included as an auxiliary task. This baseline tests whether simply adding all remaining genes can improve primary-gene prediction without task selection.
3) ``{\bf AGL}+ Uncertainty''~\cite{kendall2018multi}: Instead of selecting a subset of auxiliaries, this variant keeps all auxiliary genes but assigns each task a weight that is learned from uncertainty: tasks with lower predictive uncertainty receive higher weights, while highly uncertain tasks are down-weighted.
4) ``{\bf AGL}+ AuxLearn''~\cite{navon2020auxiliary}: This variant retains all auxiliary genes and combines their losses through \emph{AuxLearn}, which feeds the individual losses into a small neural network that predicts a non-linear weighting for each task according to its estimated contribution to the target task.
5) ``{\bf AGL}+ AMAL'':  Based on the adaptive multi-task weighting strategy AMAL~\cite{sivasubramanian2023adaptive}, this variant attaches one learnable scalar to every auxiliary task and multiplies that scalar by the task’s loss.  A temperature-controlled sigmoid keeps each scalar in the interval \([0,1]\); values close to~1 retain the task, whereas values near~0 effectively drop it, so the mechanism performs soft task selection driven by the contribution of each auxiliary gene to the primary loss.
6) ``{\bf AGL}+{\bf DkGSB}(ours)'': Uses the proposed differentiable top-$k$ mask on the HVG ranking; $k$ is learned via the bi-level scheme in Section 4.3.



For a fair comparison, we adopt ST-Net \cite{he2020integrating}, which is the most widely used architecture for gene-expression prediction, as the common backbone in all experiments.
Our framework is model-agnostic and could be applied to any ST predictor, but fixing the backbone isolates the
effect of the auxiliary-loss strategies. Apart from the way auxiliary losses are combined, every competing method shares the same network architecture, optimization schedule, and hyper-parameters used in AGL.

\noindent
{\bf Intra-batch experimentu.}
{\bf BOWEL A}, {\bf BOWEL B}, and {\bf OVARY} in Table~\ref{tab:comparison_intra} presents the PCC scores for the three intra-batch tissues. Training on the $50$ primary genes alone (``PGL'') yields the lowest accuracy throughout. 
Adding all auxiliary genes without selection (``\textbf{AGL}~+~All'') improves performance across all datasets, indicating that auxiliary signals are broadly beneficial.
 Weighting losses by predictive uncertainty (``\textbf{AGL}~+~Uncertainty'')
 yields only a modest performance gain, while the adaptive weighting schemes ``\textbf{AGL}~+~AuxLearn'' and ``\textbf{AGL}~+~AMAL'' offer no consistent improvement: optimizing weights for nearly $20{,}000$ tasks proves difficult without additional guidance. 
The proposed ``\textbf{AGL}+\textbf{DkGSB}'', which replaces exhaustive subset search with a single learnable HVG-based cut-off, achieves superior performance over all other methods across all tissue types.
 

\noindent
{\bf Inter-batch experiment.}
In the inter-batch experiment on the \textbf{HEART} dataset in Table \ref{tab:comparison_intra}, all methods showed decreased performance compared to those in the intra-batch datasets. This result also suggests that batch effects have a substantial impact on the performance of spatial gene expression prediction.
Even under this setting, applying ``{\bf AGL}'' successfully outperforms ``PGL'', and further performance improvement is achieved by selecting auxiliary genes using the proposed ``{\bf DkGSB}''.
This result confirms that our approach retains its effectiveness even in the presence of batch effects.

\begin{figure}
      \centering
        \includegraphics[width=0.7\linewidth]{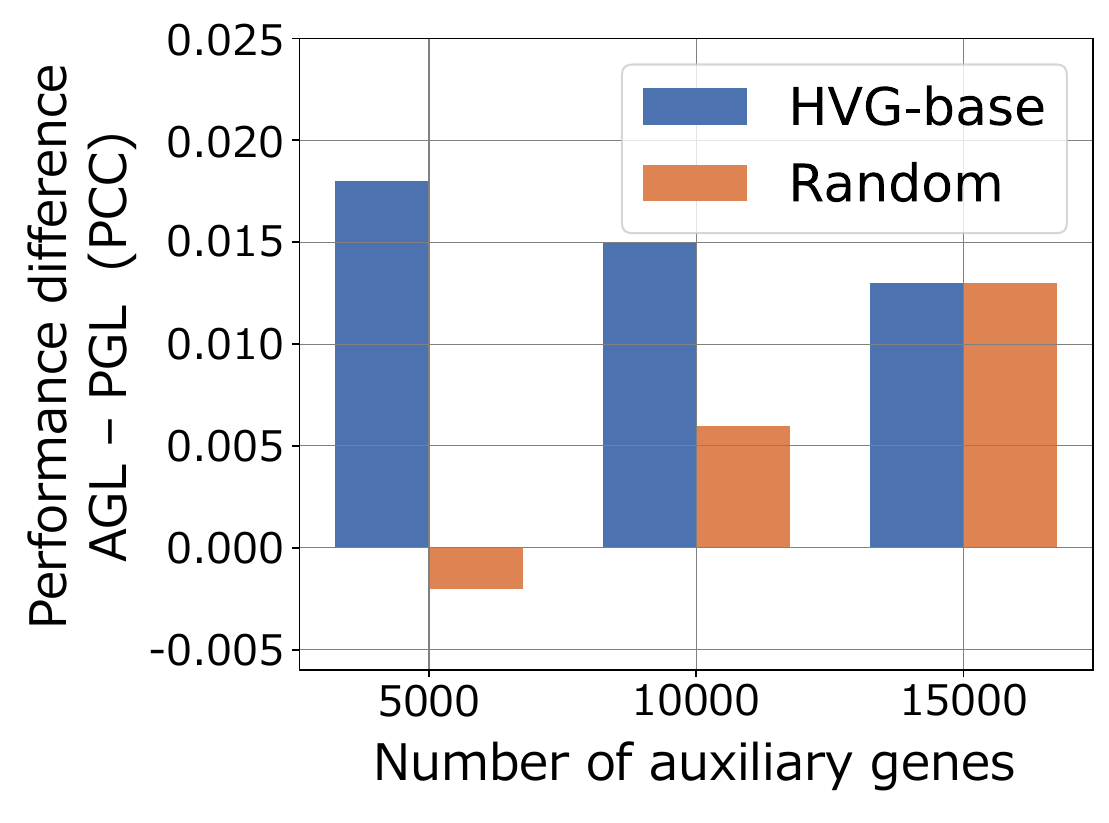}
        \vspace{-2mm}
        \caption{{\bf Reasonability of HVG score-based selection.} 
        Performance of primary gene expression estimation using HVG score–based selection (blue) and random selection (orange).
        The vertical axis shows the performance difference between models trained with auxiliary genes selected by each method and those trained using only primary genes ("PGL"), while the horizontal axis indicates the number of selected auxiliary genes.
        The experiments were conducted using  { \bf BOWEL B}.
        }
        \label{fig:preliminary}
        \vspace{-2mm}
\end{figure}

\subsection{Analysis}
\noindent
{\bf Reasonability of HVG score-based selection.}
To demonstrate the reasonability of selecting auxiliary genes based on HVG scores, we compared the performance of primary gene estimation using HVG score-based selection and random selection, with 5{,}000, 10{,}000, and 15{,}000 auxiliary genes, respectively.

Figure~\ref{fig:preliminary} shows, on the vertical axis, the performance difference between models trained with a subset of selected auxiliary genes and those trained using only primary genes (``PGL''), and on the horizontal axis, the number of selected auxiliary genes. 
Values greater than 0.0 indicate an improvement in performance, while values below 0.0 represent a decrease in performance.
The blue bars represent the HVG score-based selection method, while the orange bars represent the random selection method.
The experiments were conducted using {\bf BOWEL B}.


Experimental results show that when the number of auxiliary genes is 5{,}000 or 10{,}000, the HVG score-based selection yields a significantly greater performance improvement compared to the random selection method. In particular, at 5{,}000 auxiliary genes, the random selection method even leads to a decrease in performance.
When the number of auxiliary genes approaches the total number of genes (15{,}000), both methods yield similar levels of performance improvement.
These results suggest that the HVG score-based selection method is capable of preferentially selecting auxiliary genes that contribute to improving the prediction performance of the target genes.

\noindent

\begin{figure}
      \centering
        \includegraphics[width=1.0\linewidth]{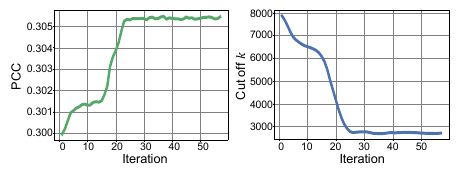}
        \vspace{-7mm}
        \caption{{\bf Behavior during cut-off $k$ optimization.} The left panel shows the changes in validation performance during the optimization process of the $outer$ loop, while the right panel shows the changes in the cut-off $k$. The experiments were conducted using  { \bf HEART} dataset.
        }
        \label{fig:meta_behavior1}
\end{figure}

{
\renewcommand{\arraystretch}{1.2}
\begin{table*}[t]
    \def\@captype{table}
      \makeatother
        \centering
        \caption{{\bf Robustness to the number of primary genes.}  The performance of ``PGL'', ``{\bf AGL} + All'', and ``{\bf AGL} + {\bf DkGSB}'' when the number of primary genes is varied among 25, 50, 75, and 100. The performance of each method is evaluated via cross-validation on the {\bf BOWEL A} dataset. Reported values indicate the mean and standard deviation, with the best performances highlighted in bold.}
        \vspace{-2mm}
        \scalebox{1.0}{
     \begin{tabular}{cc||c|c|c|c||c} \hline
         \multicolumn{2}{c||}{Method} &{\bf 25} &{\bf 50}&{\bf 75} &{\bf 100}&{\bf Average}   \\ \hline 
                PGL &&0.445$\pm$0.009&0.514$\pm$0.009&0.542$\pm$0.008&0.561$\pm$0.009&0.516\\
                {\bf AGL} &$+$ ALL&0.481$\pm$0.009&0.542$\pm$0.008&0.557$\pm$0.007&0.571$\pm$0.008&0.538\\
        \rowcolor{gray!15} {\bf AGL}& $+$ {\bf DkGSB} &{\bf 0.487}$\pm$0.008&{\bf 0.551}$\pm$0.009&{\bf 0.564}$\pm$0.008&{\bf 0.579}$\pm$0.008&{\bf 0.545}\\ \hline 


        \hline  
        \end{tabular}
        }
        \vspace{-1mm}
        \label{tab:robustness}
\end{table*}
}

\noindent
{\bf Behavior during cut-off $k$ optimization.}
Figure~\ref{fig:meta_behavior1} illustrates how the bi-level algorithm optimizes the cut-off $k$ on the \textbf{HEART} dataset. 
The \emph{left graph} traces the validation Pearson correlation (PCC) of the primary genes over successive outer iterations. Beginning with the full mask ($k=n_{\mathrm{aux}}$), the PCC rises monotonically for about 15–20 outer steps and then stabilises at $\approx 0.305$.
The \emph{right graph} shows the simultaneous evolution of the cut-off $k$. Over the same 20 iterations the algorithm lowers $k$ from the full
auxiliary pool to $k = 2{,}698$, after which both $k$ and the PCC remain essentially flat. Since the \textbf{HEART} dataset contains roughly
15{,}000 auxiliary genes, the final threshold retains only $\sim 18\%$ of the candidates while discarding the remaining $82\%$. The fact that the accuracy curve has already reached its plateau when $k$ converges confirms that the optimizer has settled on a subset that is near-optimal for this tissue.

\begin{figure}
      \centering
        \includegraphics[width=1.0\linewidth]{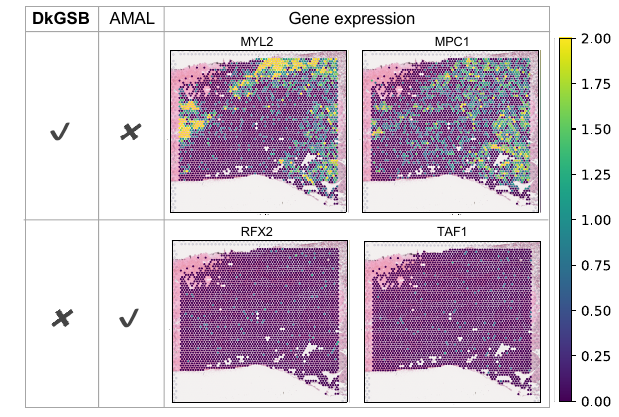}
        \vspace{-5mm}
        \caption{{\bf Visualization of the expression levels for the selected auxiliary genes.} In the top row, the expression patterns of genes that were selected by the proposed method but not by ``{\bf AGL}+AMAL,'' and in the bottom row, the opposite: genes that were not selected by the proposed method but were selected by ``{\bf AGL}+AMAL.'' The name of the visualized gene is shown at the top of each slide.
        }
        \label{fig:selected_gene}
        \vspace{-3mm}
\end{figure}

\noindent
{\bf Visualization of the expression levels for the selected auxiliary genes.}
To compare the expression patterns of auxiliary genes selected by the proposed ``{\bf DkGSB}'' and the conventional auxiliary task learning method ``\textbf{AGL}~+AMAL'', Figure~\ref{fig:selected_gene} visualizes the expression of the selected genes on the {\bf HEART} dataset.
The top row shows genes selected by ``{\bf AGL}+{\bf DkGSB}'' but not by ``{\bf AGL}+AMAL,'' and the bottom row shows the reverse.
The name of each visualized gene appears above its slide.

As shown in the top row, the proposed ``{\bf AGL}+{\bf DkGSB}'' tends to select highly expressed genes as auxiliary genes, whereas, as shown in the bottom row, ``{\bf AGL}+AMAL'' selects genes that are barely expressed across the entire slide and are unlikely to contribute effectively to the learning process.
This result suggests that ``{\bf AGL}+AMAL'' is unable to effectively select appropriate gene combinations because it directly tackles a high-dimensional combinatorial optimization problem.

\noindent
{\bf Robustness to the number of primary genes.}
Although the Hest-1k dataset is commonly used with the top 50 highly variable genes as prediction targets~\cite{jaume2024hest, cho2025mv_}, whose expression levels consistently rise above the noise floor and are considered reliable, we conducted experiments by varying the number of primary genes to demonstrate the robustness of the proposed ``{\bf AGL}+{\bf DkGSB}'' to such changes.
Table~\ref{tab:robustness} presents the performance of ``PGL'', ``{\bf AGL} + All'', and ``{\bf AGL} + {\bf DkGSB}'' on the {\bf BOWEL A} dataset when the number of primary genes is set to 25, 50, 75, and 100.
These results indicate that even when the number of primary genes varies, leveraging previously discarded genes as auxiliary genes remains effective.
Moreover, the proposed ``{\bf DkGSB}'' consistently improves performance regardless of the number of primary genes. These findings demonstrate that the proposed ``\textbf{AGL}+\textbf{DkGSB}'' is robust to variations in the number of primary genes.


\section{Limitations}
While formulating the auxiliary gene selection problem as a top‑$k$ choice based on HVG score ranking improves primary gene prediction, several limitations remain.
First, since the approach relies solely on HVG scores, it is challenging to fully capture the underlying biological functional relationships among genes. When incorporating the biological relationships, the problem involves both determining the set of informative genes to select and identifying which biological functions are activated, and it is complicated. In the present study, we restricted our analysis to the informativeness of genes and did not model pathway activation effects as a first step.
As a potential direction for future work, one could consider predicting the contribution of each gene based on both the strength of its expression signal and its functional relevance to the primary genes, and using this information to guide auxiliary gene selection.


Second, the contribution of an auxiliary gene to the estimation of primary genes may vary depending on the spatial location within the tissue; a gene may be informative in one region but act as noise in another.
A spatially adaptive mechanism that detects and down-weights unhelpful image–gene pairs during training, for example by monitoring per-sample losses or employing co-teaching strategies~\cite{han2018co} from the noisy-label literature, could further improve robustness.  
In such a case, the addition of data selection alongside gene selection would increase the overall complexity of the problem, potentially requiring the development of an alternative framework.



\section{Conclusion}
In this study, we proposed \textit{Auxiliary Gene Learning} (AGL), which leverages the benefits of previously ignored genes by reformulating their expression estimation as auxiliary tasks and jointly training them with the primary tasks.
In AGL, it is necessary to select an appropriate subset of auxiliary genes from a larger set that is often affected by observational noise. However, this becomes a high-dimensional combinatorial optimization problem, making it challenging to solve.
To overcome this challenge, we proposed Prior-Knowledge-Based Differentiable Top-$k$ Gene Selection via Bi-level Optimization (DkGSB), which ranks all auxiliary gene candidates and performs top-$k$ selection in a differentiable manner.
The experimental results demonstrate the effectiveness of incorporating previously ignored genes into the learning process as auxiliary tasks, and show that the proposed DkGSB method outperforms conventional auxiliary task learning approaches.

\section{Acknowledgments}
This work was supported by JPMJBS2406, SIP-JPJ012425,
ASPIRE Grant Number JPMJAP2403, JSPS KAKEN JP24KJ2205, JP23K18509, JP25K22846, and JP23KJ1723;
JST ACT-X JPMJAX23CR; the AMED grant 24ama221609h0001 (P-PROMOTE) (to YK);
and the National Cancer Center Research and Development Fund 2024-A-6 (to YK).


\bibliography{aaai2026}

\end{document}